\documentclass[10pt,conference]{IEEEtran}
\IEEEoverridecommandlockouts
\usepackage{cite}
\usepackage{amsmath,amssymb,amsfonts}
\usepackage{algorithmic}
\usepackage{graphicx}
\usepackage{textcomp}
\usepackage{xcolor}
\def\BibTeX{{\rm B\kern-.05em{\sc i\kern-.025em b}\kern-.08em
    T\kern-.1667em\lower.7ex\hbox{E}\kern-.125emX}}

\usepackage[compatibility=false]{caption}
\usepackage[subrefformat=parens]{subcaption}
\usepackage{algorithm}
\usepackage{booktabs}
\usepackage{multirow}
\usepackage{comment}

\begin{document}

\title{{Multi-Objective Multi-Agent Path Finding Based on Two-Stage Evolutionary Algorithms}\thanks{This work was supported by JSPS Grant-in-Aid for Scientific Research (B) (Grant Number: 24K03001).}}

\author{\IEEEauthorblockN{Haruto Ando\IEEEauthorrefmark{1}, Yoshiki Nogata\IEEEauthorrefmark{2},
Tomohiro Harada\IEEEauthorrefmark{2},
Fumito Uwano\IEEEauthorrefmark{3}}
\IEEEauthorblockA{\IEEEauthorrefmark{1}Faculty of Engineering, Saitama University, Saitama, Japan \\
Email: ando.h.785@ms.saitama-u.ac.jp}
\IEEEauthorblockA{\IEEEauthorrefmark{2}Graduate School of Science and Engineering, Saitama University, Saitama, Japan \\
Email: \{nogata.y.564@ms., tharada@mail.\}saitama-u.ac.jp}
\IEEEauthorblockA{\IEEEauthorrefmark{3}Faculty of Environmental, Life, Natural Science and Technology, Okayama University, Okayama, Japan\\
Email: uwano@okayama-u.ac.jp}}

\maketitle

\begin{abstract}
Multi-agent path finding (MAPF) is a critical problem in applications such as warehouse automation and autonomous driving. 
In real-world settings, multiple conflicting objectives, such as minimizing path length and ensuring safety, often must be optimized simultaneously, motivating the study of multi-objective MAPF (MOMAPF).
However, existing MOMAPF approaches suffer from severe scalability limitations as the number of agents and the environment size increase. 
To address this challenge, we propose Two-Stage Evolutionary MOMAPF (TSE-MOMAPF), which optimizes a single agent's path via NSGA-II and refines the path set using an SMS-EMOA-based framework.
Experiments on various grid maps show that the proposed method achieves superior scalability and provides broader Pareto front coverage in medium- and large-scale scenarios compared to an existing method. 
These results indicate that the proposed evolutionary framework effectively balances solution diversity and computational efficiency under strict time constraints.
\end{abstract}


\section{Introduction}
Multi-agent path finding (MAPF) is the problem of obtaining a set of collision-free paths that connect the respective start and goal positions of multiple agents~\cite{Alkazzi_2024,Stern_2021}. 
MAPF has attracted significant attention due to its wide range of applications, including warehouse automation using transport robots~\cite{Guizzo_2008} and autonomous vehicle driving~\cite{dresner_2008}.

In real-world settings, MAPF often involves multiple conflicting objectives, such as minimizing path length and ensuring path safety. 
This study, therefore, focuses on multi-objective MAPF (MOMAPF)~\cite{weise_2020}, which generalizes MAPF to handle multiple objectives. 
A notable approach for MOMAPF is multi-objective conflict-based search (MO-CBS)~\cite{Ren_2023}, which extends the conflict-based search (CBS) framework~\cite{sharon_2015} from single-objective to a multi-objective optimization.
However, it suffers from exponential growth in the search space, hindering efficiency in large-scale environments.

To address this challenge, we propose a novel method for solving MOMAPF using multi-objective evolutionary algorithms (MOEAs)~\cite{Sharma2022-qi}. 
Our method aims to overcome the inefficiencies of conventional approaches and to discover high-quality Pareto solutions within a realistic running time. 

To investigate the effectiveness of the proposed method, we conduct experiments comparing it with MO-CBS under identical time constraints across four grid-based maps of varying sizes and numbers of agents.

\section{Problem Formulation}
In MOMAPF, we define $A=\left\{1, 2, \dots, N\right\}$ as a set of $N$ agents placed in a 2D space. 
This 2D space is defined as a grid map $G$ where cells are connected in four directions. 

Each agent $i \in A$ has a start position $s^i$ and a goal position $g^i$, and all agents start moving at time $t=0$. 
An agent can move to one of the four neighboring grid cells in each time unit. 
$\pi^i\left(i \in \left\{1, 2, \dots, N\right\}\right)$ represents the path for agent $i$ from $s^i$ to $g^i$ expressed as $\pi^i=\left\{s^i, p^i_1, p^i_2, \dots, g^i\right\}$, where $p^i_k$ represents the coordinates of agent $i$ at time $k$. 
The set of paths for all agents, $\pi=\left\{\pi^1, \pi^2, \dots, \pi^N\right\}$, is one solution for MOMAPF.

\subsection{Constraints}
A solution $\pi$ must satisfy the constraint that no conflicts occur between any pair of agents $i \neq j$. Solutions with one or more conflicts are regarded as infeasible.
Conflicts are defined by two patterns: vertex conflicts and edge conflicts. 
A vertex conflict occurs when two agents occupy the same coordinate $p$ at the same time $t$. 
An edge conflict occurs when two agents pass through the same edge from opposite directions between time $t$ and $t+1$. 

\subsection{Objectives}
The MOMAPF search considers $M$ different objectives. 
Each objective is defined using a function $f_m\left(m \in \left\{1, 2, \dots, M\right\}\right)$, and the multi-objective function $F$ to be minimized is defined as:
\begin{equation}
\text{minimize} \: F(\pi)=\left(f_1\left(\pi\right), f_2\left(\pi\right), \dots, f_M\left(\pi\right)\right)
\label{eq2}
\end{equation}

Since MOMAPF is a multi-objective optimization problem, solution dominance is defined by comparing their corresponding objective function vectors. 
Given two solutions $\pi_a$ and $\pi_b$, along with their objective function vectors $F\left(\pi_a\right)$ and $F\left(\pi_b\right)$, $\pi_a$ is defined to dominate $\pi_b$ (denoted as $\pi_a \prec \pi_b$) if $f_m\left(\pi_a\right) \le f_m\left(\pi_b\right)$ for all objectives and $f_m\left(\pi_a\right) < f_m\left(\pi_b\right)$ for at least one objective. 
Conversely, if two solutions do not dominate each other, they are defined as non-dominated. 
The Pareto optimal solution set (POS) refers to the set of non-dominated solutions that are not dominated by any other solution among all feasible solutions $\Lambda$, which is defined as:
\begin{equation}
POS=\{\pi^*\in \Lambda \mid \forall\: \pi^\prime\in \Lambda, \pi^\prime\not\prec\pi^*\}
\label{eq3}
\end{equation}

In the context of MOMAPF, the goal is to compute a diverse set of collision-free (i.e., feasible) Pareto-optimal path solutions within realistic time constraints.

While optimizing a single agent's path is common, MAPF requires solving this problem for multiple agents and then detecting and resolving conflicts. Furthermore, Pareto-optimal solutions must be considered at both the single- and multi-agent optimization levels. These characteristics make MOMAPF challenging for EAs or heuristics.

\section{Previous Work: MO-CBS}
MO-CBS~\cite{Ren_2023} has been proposed as a conventional method for MOMAPF. 
MO-CBS is an algorithm that extends the CBS framework~\cite{sharon_2015}, which addresses single-objective MAPF, to multi-objective optimization. 
Similar to CBS, it is characterized by a two-stage search process: a low-level search and a high-level search. 
MO-CBS processes are described below.

\subsection{Low-Level Search}
In the low-level search, path planning is conducted for a single agent $i$. 
This path planning is subject to a constraint set $\Omega^i$ provided by the high-level search to avoid conflicts with other agents. 
From the paths that satisfy all constraints, a Pareto optimal path set $\Pi^i_*=\left\{\pi^i_a, \pi^i_b, \dots\right\}$ is calculated. 
For path generation, an existing multi-objective path planning algorithm for a single agent, such as NAMOA*-dr~\cite{pulido_2015}, is used. 
In particular, for two objectives, BOA*~\cite{ulloa_2020}, an expedited version of NAMOA*-dr, is used.

\subsection{High-Level Search}
In the high-level search, a tree structure is used to resolve conflicts among agents using the CBS framework. 
Each node in the tree consists of $P=\left\{\pi, F(\pi), \Omega\right\}$. 
$F(\pi)$ represents the objective function value vector of solution $\pi$, and $\Omega$ represents the set of constraints imposed on this node. 
Each node is added to a priority queue $OPEN$, which is sorted lexicographically based on mutually non-dominated objective function value vectors $F(\pi)$. 
When executing the search, one node $P$ is selected from the queue. 
If a conflict is found in the solution $\pi$ of the selected node, constraints are generated for each agent to avoid the conflict. 
After splitting the node, a low-level search is performed on each resulting node to update $\pi$. 
If no conflict exists in the solution, the solution $\pi$ and its objective function value vector $F(\pi)$ are added to the solution set $\mathcal{C}$.

\subsection{Algorithm of MO-CBS}
In MO-CBS, the initial path set $\Pi^i_0$ for agent $i$ is generated using the algorithms employed in the low-level search (NAMOA*-dr or BOA*). 
An initial solution set $\Pi_0=\left\{\pi_0|\pi_0=\left(\pi^1_0, \pi^2_0, \dots, \pi^N_0\right),\pi^i_0\in\Pi^i_0\right\}$ is then created by considering all possible combinations of paths from the initial path sets of all agents. 
For each of these solutions $\pi_0\in\Pi_0$, a node $P$ is generated and added to $OPEN$. 
This node contains the corresponding objective function value vector $F(\pi)$ and an empty constraint set $\Omega=\emptyset$.

After initializing $OPEN$, the high-level and low-level searches are repeated alternately until $OPEN$ becomes empty. 
When adding solution $\pi$ and its objective function value vector $F(\pi)$ to the solution set $\mathcal{C}$ during the high-level search, any objective function value vectors within $\mathcal{C}$ that are dominated by $F(\pi)$ are removed. 
Consequently, only non-dominated solutions remain in the solution set $\mathcal{C}$ at the end of the search. 
Conversely, if a solution that dominates the objective function vector $F(\pi)$ already exists in the solution set $\mathcal{C}$, it is not added. 
This ensures that the solution set $\mathcal{C}$ becomes the Pareto optimal solution set upon completion of the search.

\section{Proposed Method}
MO-CBS becomes inefficient as the problem scale increases, for example, in the number of agents or the environment size. 
To address this issue, this study proposes a novel approach for solving MOMAPF, named Two-Stage Evolutionary MOMAPF (TSE-MOMAPF), that ensures scalability for large-scale problems by incorporating MOEAs. 

The proposed method consists of two stages: the first optimizes the path of a single agent, $\pi^i$, and the second optimizes the set of paths, $\pi$.
This approach is motivated by the complexity of simultaneously optimizing the paths of all agents.
Instead, our method refines the path set $\pi$ by modifying the path of a single agent, $\pi^i$, in a manner similar to a mutation process.
The processes of the algorithm are described below.

\subsection{Initialization}
At the beginning of TSE-MOMAPF, each agent's path set, $\Pi^i$, is initialized. 
The initialization process is shown in Alg.~\ref{alg_init}. 

First, the minimum-cost path for each objective $m$ is computed backward from the goal point $g^i$ using Dijkstra's algorithm to determine the minimum-cost path from any location on the map to $g^i$. 
The map area used for this search is restricted to a rectangle with vertices at the start point $s^i$ and the goal point $g^i$ of agent $i$, expanded by a fixed distance $D$. 
This improves the efficiency of the initialization process for large-scale problems.

Subsequently, a path is constructed from the start point $s^i$ to the goal point $g^i$ by following the minimum cost path $\pi^i_m$.
During this process, randomness is introduced by selecting paths other than the minimum cost path, including backtracking paths, with a constant probability $P_{dev}$. 
This approach is inspired by the algorithm proposed by Eppstein~\cite{eppstein_1998} and ensures both path quality and randomness. 

For each objective, either $\left\lfloor{N_{s}/M}\right\rfloor$ or $\left\lceil{N_{s}/M}\right\rceil$ paths are generated such that $\left|\Pi^i\right| = N_{s}$. 
The decision of which number of paths to generate for each objective is made randomly.

\begin{algorithm}[!tb]
\caption{Pseudo-code of single agent path set initialization}
\begin{algorithmic}[1]
\REQUIRE $s^i$, $g^i$, $D$, $P_{dev}$
\ENSURE $\Pi^i$
\STATE $G^\prime \leftarrow$ Area in $G$ limited by $s^i$, $g^i$, and $D$
\FOR{Each objective $m$}
\STATE Calculate minimum cost path from $v^\prime \in G^\prime$ to $g^i$
\STATE $\pi^i_m \leftarrow$ Minimum cost path from $s^i$ to $g^i$
\FOR{$\left\lfloor{N_{s}/M}\right\rfloor$ or $\left\lceil{N_{s}/M}\right\rceil$}
\STATE $\pi^i \leftarrow$ Deviate from $\pi^i_m$ with $P_{dev}$
\STATE $\Pi^i \leftarrow \Pi^i \cup \{\pi^i\}$
\ENDFOR
\ENDFOR
\RETURN $\Pi^i$
\end{algorithmic}
\label{alg_init}
\end{algorithm}

\subsection{Optimizing Single Agent Path}
NSGA-II, a widely used MOEA for bi- or three-objective optimizations, is employed to optimize the path $\pi^i$ of a single agent.
This single-agent optimization is employed during initial solution generation and as a mutation operator to optimize multi-agent path sets (as described in Section~\ref{sec_ma}).

\subsubsection{Crossover}
The crossover between two paths depends on the number of locations at which agents pass through the same coordinate $p$ at the same time $t$. 
Fig.~\ref{fig_tpcx} through Fig.~\ref{fig_pscx} illustrate these crossover methods: the start and end points of the paths are represented by circles and squares, respectively, while the aforementioned locations are indicated by star symbols. 
If there are two or more shared locations, two points are randomly selected, and a two-point crossover is applied as shown in Fig.~\ref{fig_tpcx}. 
If there is only one shared location, a one-point crossover is used as shown in Fig.~\ref{fig_opcx}. 
If there are no shared locations, one random point is chosen from each path, and a path synthesis crossover is performed to connect these two points as shown in Fig.~\ref{fig_pscx}. 
The connecting path is generated to minimize the cost of a randomly selected single objective. Fig.~\ref{fig_pscx} shows the resulting path when the path length is minimized.

\subsubsection{Mutation}
\begin{figure}[!tb]
\centering
\begin{minipage}[b]{0.2\textwidth}
\centering
\includegraphics[width=\linewidth]{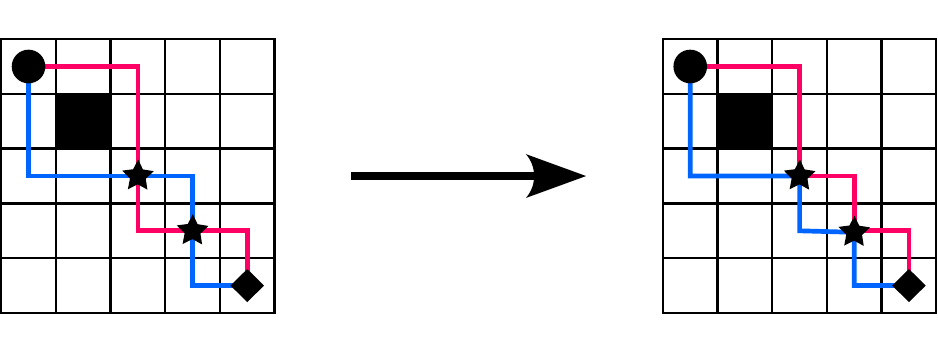}
\subcaption{Two point crossover}
\label{fig_tpcx}
\end{minipage}
\hfill
\begin{minipage}[b]{0.2\textwidth}
\centering
\includegraphics[width=\linewidth]{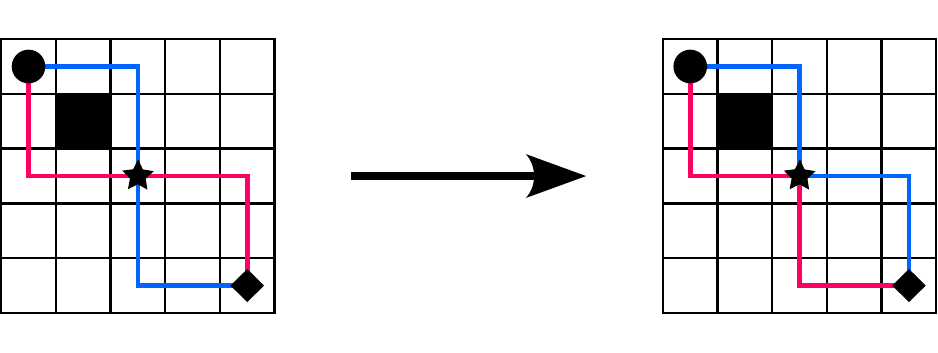}
\subcaption{One point crossover}
\label{fig_opcx}
\end{minipage}
\hfill
\begin{minipage}[b]{0.2\textwidth}
\centering
\includegraphics[width=\linewidth]{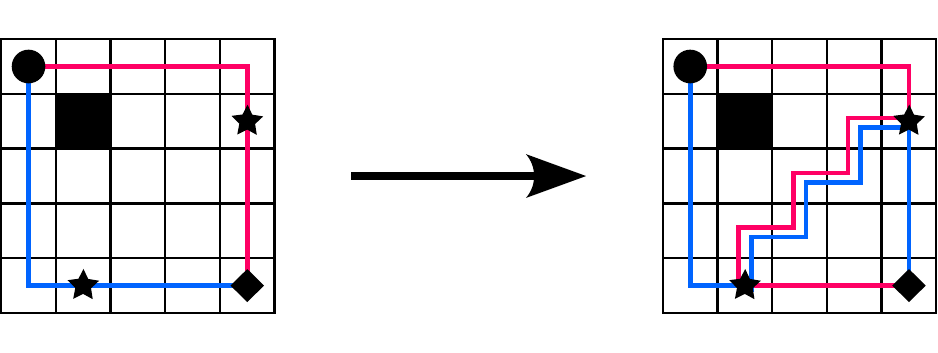}
\subcaption{Path synthesis crossover}
\label{fig_pscx}
\end{minipage}
\hfill
\begin{minipage}[b]{0.2\textwidth}
\centering
\includegraphics[width=\linewidth]{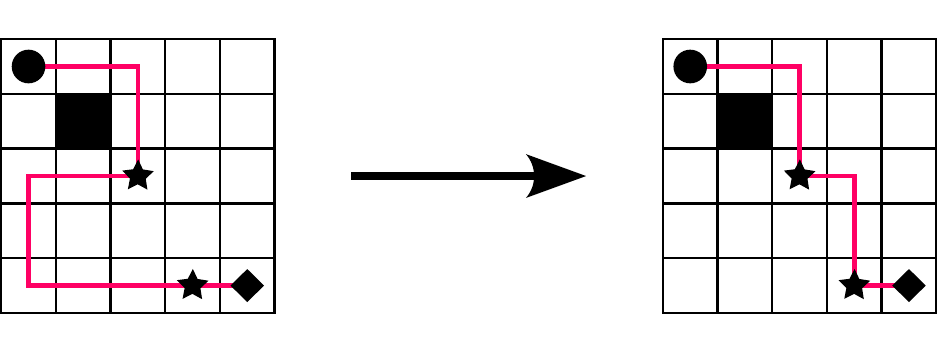}
\subcaption{An example of mutation}
\label{fig_mut}
\end{minipage}
\caption{Examples of crossovers and mutation}
\label{fig_cx_mut}
\end{figure}

The mutation involves selecting two random points in the path and mutating the segment between them. 
The new path segment is generated to minimize the cost of a randomly selected single objective. 
Fig.~\ref{fig_mut} illustrates an example of the mutation, where two randomly selected points are indicated by star symbols and connected so that the path length is minimized.

\subsubsection{Algorithm of Single Agent Optimization}
Alg.~\ref{alg_sa} shows the optimization algorithm for a single agent path $\pi^i$. 
The optimization loop from line 2 to line 5 is repeated for a specified number of generations, as per the NSGA-II framework. 
For the offspring generation in line 3, crossover and mutation are employed. 
As in NSGA-II, both parent selection in line 3 and survival selection in line 4 use non-dominated sorting and crowding distance. 
Finally, in line 7, non-dominated solutions are selected from the final population $\mathcal{P}$ to form the algorithm's output solution set $\mathcal{S}^i$.

\begin{algorithm}[!tb]
\caption{Pseudo-code of single agent path optimization}
\label{alg_sa}
\begin{algorithmic}[1]
\REQUIRE $\Pi^i$
\ENSURE $\mathcal{S}^i, \Pi^i$
\STATE $\mathcal{P} \leftarrow \Pi^i$\COMMENT{Inherit initial path set}
\REPEAT
\STATE $\mathcal{Q} \leftarrow$ Generate offspring from $\mathcal{P}$ using crossover and mutation
\STATE $\mathcal{P} \leftarrow$ Select next population from $\mathcal{P} \cup \mathcal{Q}$ using non-dominated sorting and crowding distance 
\UNTIL{Termination condition is met}
\STATE $\Pi^i\leftarrow \mathcal{P}$
\STATE $\mathcal{S}^i \leftarrow$ Non-dominated solutions in $\mathcal{P}$
\RETURN $\mathcal{S}^i, \Pi^i$
\end{algorithmic}
\end{algorithm}

\subsection{Optimizing Multiple Agent Path Set}
\label{sec_ma}
The optimization of the path set $\pi$ is inspired by SMS-EMOA \cite{beume_2007}. 
We employ SMS-EMOA because its steady-state update structure---where a fixed-size population is maintained by incrementally incorporating new offspring---naturally fits our method, in which modifying a single agent's path yields a variable number of new candidate solutions that must be selectively merged into the population based on HV contribution.
The individual processes are described below.

\subsubsection{Initialization}
After executing Alg.~\ref{alg_init} for all agents and initializing the path sets $\Pi^i$, the solution set $\Pi=\left\{\pi_1, \pi_2, \dots, \pi_{N_{m}}\right\}$, which combines the paths of all agents, is initialized. 
Each solution $\pi_k=\left\{\pi^1_k, \pi^2_k, \dots, \pi^N_k | \pi^i_k \in \Pi^i\right\}$ is constructed by combining randomly selected paths, one for each agent, from its corresponding generated path sets $\Pi^i$. 
After selecting the paths, the conflicts among agents is checked. 
If any conflicts exist, the solution is discarded, and new combinations are generated until $\left|\Pi\right|=N_{m}$ is satisfied.

\subsubsection{Offspring Generation}
One solution $\pi_k=\left\{\pi^1_k, \pi^2_k, \dots, \pi^N_k\right\}$ is randomly selected from the population $\Pi=\left\{\pi_1, \pi_2, \dots, \pi_{N_{m}}\right\}$ and used as a parent to generate offspring. 
An agent $i \in A$ is chosen from the parent $\pi_k$, and the single agent path optimization in Alg.~\ref{alg_sa} is executed for this agent. 
To improve search efficiency, the existing path set $\Pi^i$ for agent $i$ is utilized as the initial population for this optimization. 
This process results in an updated path set $\Pi^i$ and its non-dominated set $\mathcal{S}^i$. 

Next, each path $\pi^i_{new} \in \mathcal{S}^i$ is used to create offspring $\pi_{new}=(\pi_k\setminus\{\pi_k^i\})\cup\{\pi_{new}^i\}$ by replacing the path of agent $i$. 
If $\pi_{new}$ contains any conflicts, it is excluded; 
otherwise, it is added to the candidate population $\mathcal{P}^\prime$ for the next generation.

\subsubsection{Survival Selection}
The next generation $\Pi_{new}$ is selected from the candidate population $\mathcal{P}^\prime$ such that $\left|\Pi_{new}\right|=N_{m}$. 

First, non-dominated sorting is performed on the population $\mathcal{P}^\prime$ to classify individuals into several ranks $\left\{\mathcal{F}_1, \mathcal{F}_2, \dots|\left(\pi_{(a)}, \pi_{(b)},\dots\right) \in \mathcal{F}_k\right\}$ based on Pareto optimality. 
Individuals are added to the next generation $\Pi_{new}$ starting from the best rank $\mathcal{F}_1$. 
If adding the entire population of rank $\mathcal{F}_k$ results in $\left|\Pi_{new}\right|>N_{m}$, individuals are selected from $\mathcal{F}_k$ until the condition $\left|\Pi_{new}\right|=N_{m}$ is met. 
For the population truncation, the hypervolume (HV) contribution is used as the selection criterion. 
This metric quantifies each individual's contribution to the total HV of the Pareto front. 
When calculating HV contribution, the objective function values are normalized to the range $[0, 1]$, and the reference point is set to 1.1 for all objectives. 
Individuals are added to $\Pi_{new}$ one by one, starting with the one that exhibits the largest contribution. 
The new population $\Pi_{new}$ is used as the next population $\Pi$. 
Finally, the non-dominated solution set $\mathcal{S}$ serves as the final solution of the algorithm.

\subsubsection{Algorithm of Multi-agent Optimization}
Alg.~\ref{alg_ma} shows the optimization algorithm for solution $\pi$. 
Lines 1 and 2 describe the initialization of the path set $\Pi^i$ for each agent, and the solution set $\Pi$. 
From lines 3 to 13, offspring $\pi_{new}$ having no conflicts are generated and added to the population $\mathcal{P}^\prime$. 
Lines 14 to 25 involve the survival selection using non-dominated sorting and HV contribution.

\begin{algorithm}[!tb]
\caption{Pseudo-code of multi-agent path optimization}
\begin{algorithmic}[1]
\ENSURE Optimized non-dominated solutions $\mathcal{S}$
\STATE $\Pi^i\leftarrow$Initialize each agent's path using Alg.~\ref{alg_init}
\STATE $\Pi\leftarrow$Initialize population by randomly combining $\pi^i\in\Pi^i$
\REPEAT
\STATE Choose random path set $\pi_k\in\Pi$
\STATE Choose random agent path $\pi_k^i\in\pi_k$
\STATE $\mathcal{P}^\prime \leftarrow \Pi$
\STATE $\mathcal{S}^i, \Pi^i \leftarrow$ Optimize single agent $\pi^i$ using Alg.~\ref{alg_sa} initialized with $\Pi^i$
\FORALL{$\pi^i_{new} \in \mathcal{S}^i$}
\STATE $\pi_{new} \leftarrow \left(\pi_k \setminus \left\{\pi_k^i\right\}\right) \cup \left\{\pi^i_{new}\right\}$
\IF{$\pi$ does not have conflict}
\STATE $\mathcal{P}^\prime\leftarrow  \mathcal{P}^\prime\cup \{\pi_{new}\}$
\ENDIF
\ENDFOR
\STATE $\mathcal{F}_1, \mathcal{F}_2,\dots \leftarrow$ Non-dominated sort$(\mathcal{P}^\prime)$
\STATE $\Pi_{new} \leftarrow \emptyset$
\FORALL{$\mathcal{F}_i \in \left\{\mathcal{F}_1, \mathcal{F}_2,\dots \right\}$}
\IF {$\left|\mathcal{F}_i \cup \Pi_{new}\right| \leq N_{m}$}
\STATE $\Pi_{new} \leftarrow \mathcal{F}_i \cup \Pi_{new}$
\ELSE
\STATE Calculate HV contributions for all solutions in $\mathcal{F}_i$
\STATE $\mathcal{F}_i^\prime \leftarrow$ Select solutions having top $N_m-\left|\Pi_{new}\right|$ HV contributions
\STATE $\Pi_{new}\leftarrow \Pi_{new}\cup \mathcal{F}_i^\prime$
\ENDIF
\ENDFOR
\STATE $\Pi \leftarrow \Pi_{new}$
\STATE $\mathcal{S} \leftarrow $ Non-dominated solutions in $\Pi \cup \mathcal{S}$
\UNTIL{Termination condition is met}
\RETURN $\mathcal{S}$
\end{algorithmic}
\label{alg_ma}
\end{algorithm}

\section{Experiment}
To evaluate the effectiveness of TSE-MOMAPF, experiments are conducted to compare it with MO-CBS. 

\subsection{Problem Settings}
The experiment focuses on MOMAPF on a grid map $G$. 
The grid maps $G$ and the start, and goal points $\left(s^i, g^i\right)$ are obtained from the ``random'' category of the existing MAPF dataset~\cite{Stern_2021}.
In this dataset, obstacles on the map are placed on each grid cell with a 10\% probability, and the start and goal points for each agent are generated by randomly selecting reachable locations on the map.
Three map sizes are used for map $G$: $32 \times 32$ grids, $64 \times 64$ grids, and $96 \times 96$ grids. 
These maps are subsequently referred to as G32, G64, and G96. 
For G32 and G64, the existing datasets are utilized. 
On the other hand, since a G96 dataset is not provided, it is generated according to the previous study~\cite{Stern_2021}. 
The number of agents $N$ is set to $N=10$ for G32 and G96, while two settings, $N=10$ and $N=15$, are considered for G64.

In this experiment, two objectives are considered: path length (Efficiency) and path cost (Safety). 
The path length is the sum of the time steps $t$ required for all agents to reach their goals, and the path cost is the total risk of the grids each agent traverses. 
The risk of each grid is calculated as the sum of the number of obstacles in the eight surrounding grids. 
This setting is identical to the previous study~\cite{Ren_2023}.

\subsection{Parameters}
The number of path sets for each agent in the first stage optimization is set to $N_s=30$, and the number of path sets for all agents in the second stage optimization is $N_m=50$. 
The search range during initialization is $D=3$, and the probability for initial path generation is $P_{dev}=0.2$. 
The crossover and mutation rates of NSGA-II are set to $0.9$ and $0.1$, respectively.

\subsection{Evaluation Criteria}
To evaluate both methods fairly, the execution time is fixed at 3,600 seconds, and the resulting final solution sets are evaluated using HV. 
For the proposed method, the HV values are the average of 5 trials, whereas a single trial is conducted for MO-CBS because it is a deterministic method.
Each objective value is normalized to the range $[0, 1]$ using the maximum and minimum values observed in the non-dominated solution sets across all generations in each trial.
Finally, the HV is calculated using a reference point of $(1.1, 1.1)$.

\section{Result}
\begin{figure}[!tb]
\centering
\begin{minipage}[b]{0.24\textwidth}
\centering
\includegraphics[width=\linewidth]{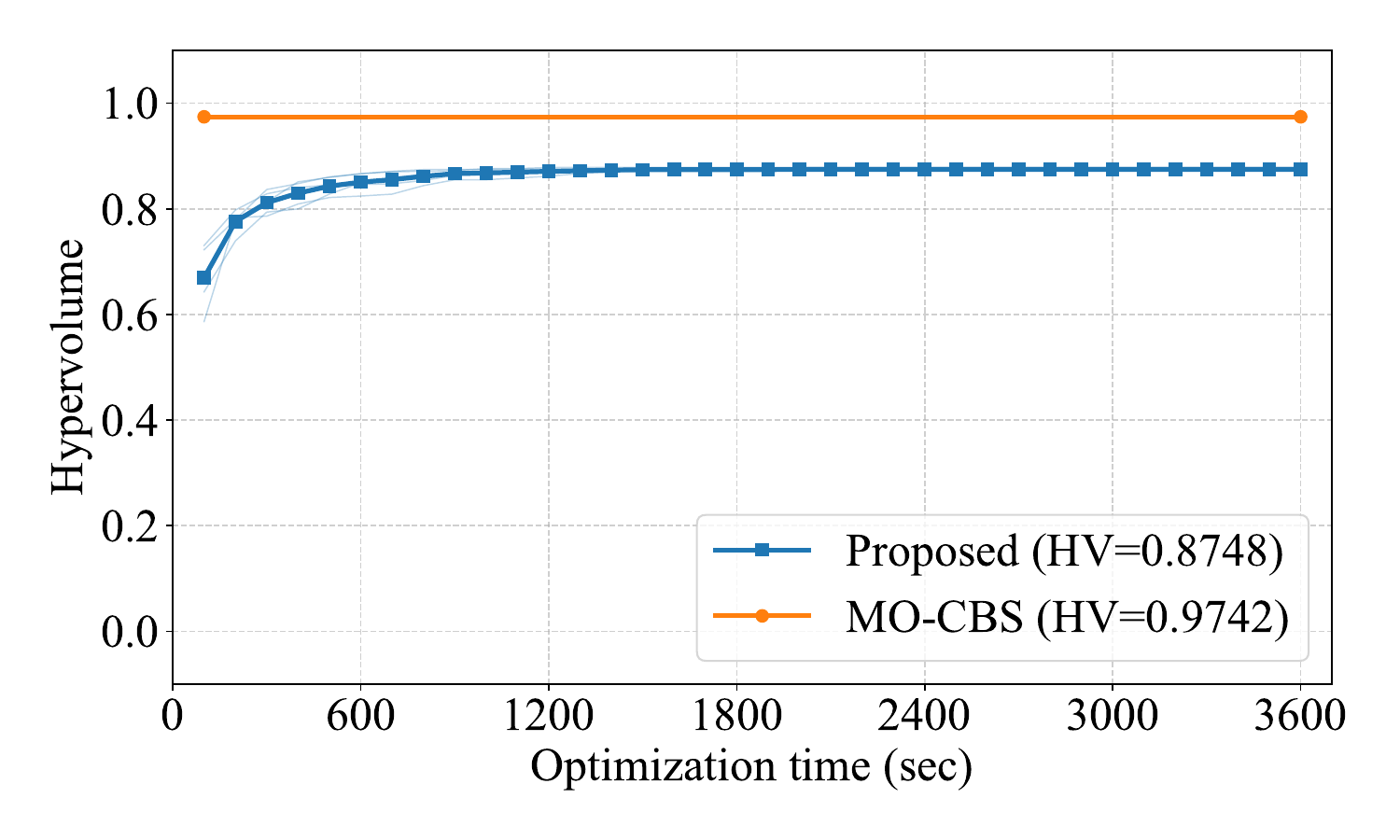}
\subcaption{G32, $N=10$}
\end{minipage}
\hfill
\begin{minipage}[b]{0.24\textwidth}
\centering
\includegraphics[width=\linewidth]{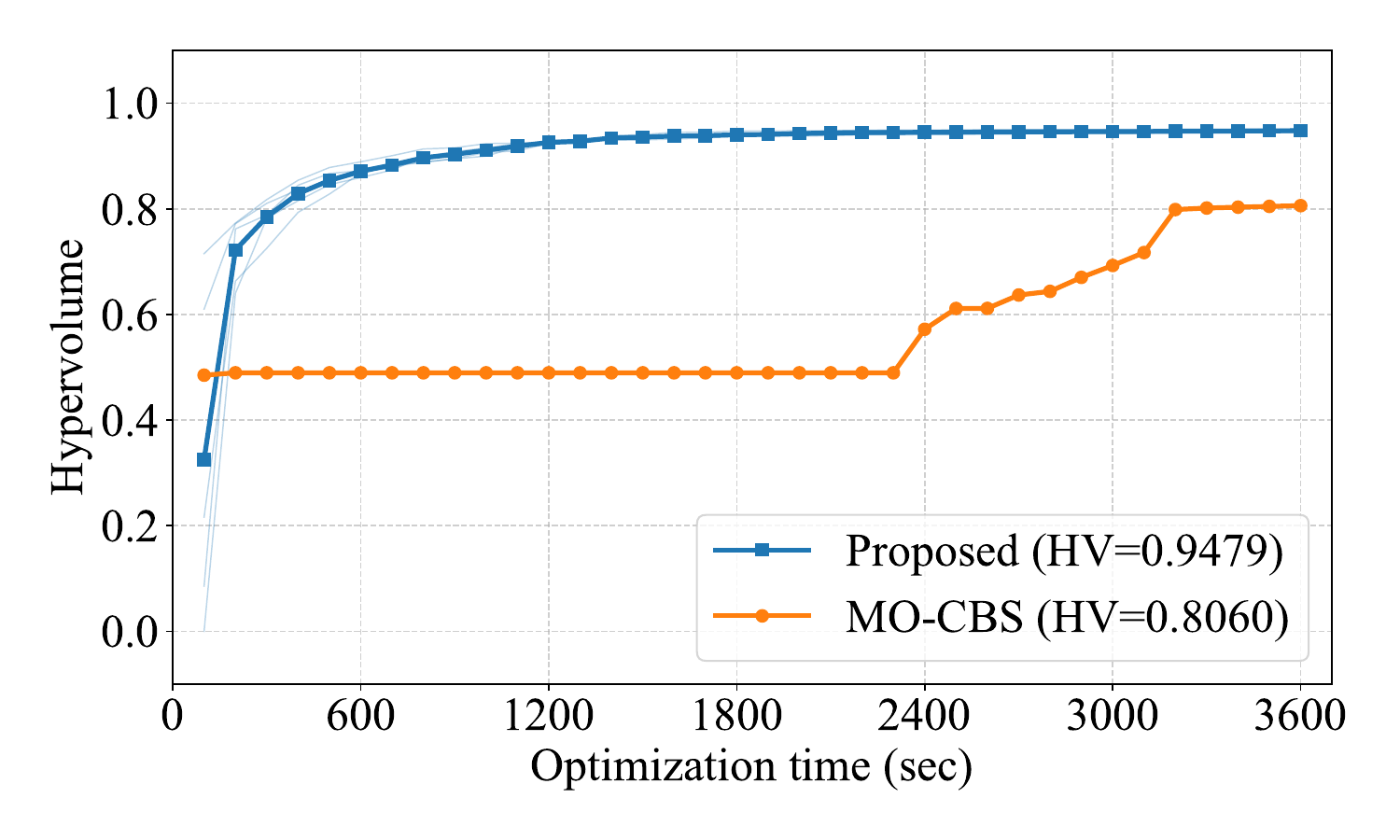}
\subcaption{G64, $N=10$}
\end{minipage}
\hfill
\begin{minipage}[b]{0.24\textwidth}
\centering
\includegraphics[width=\linewidth]{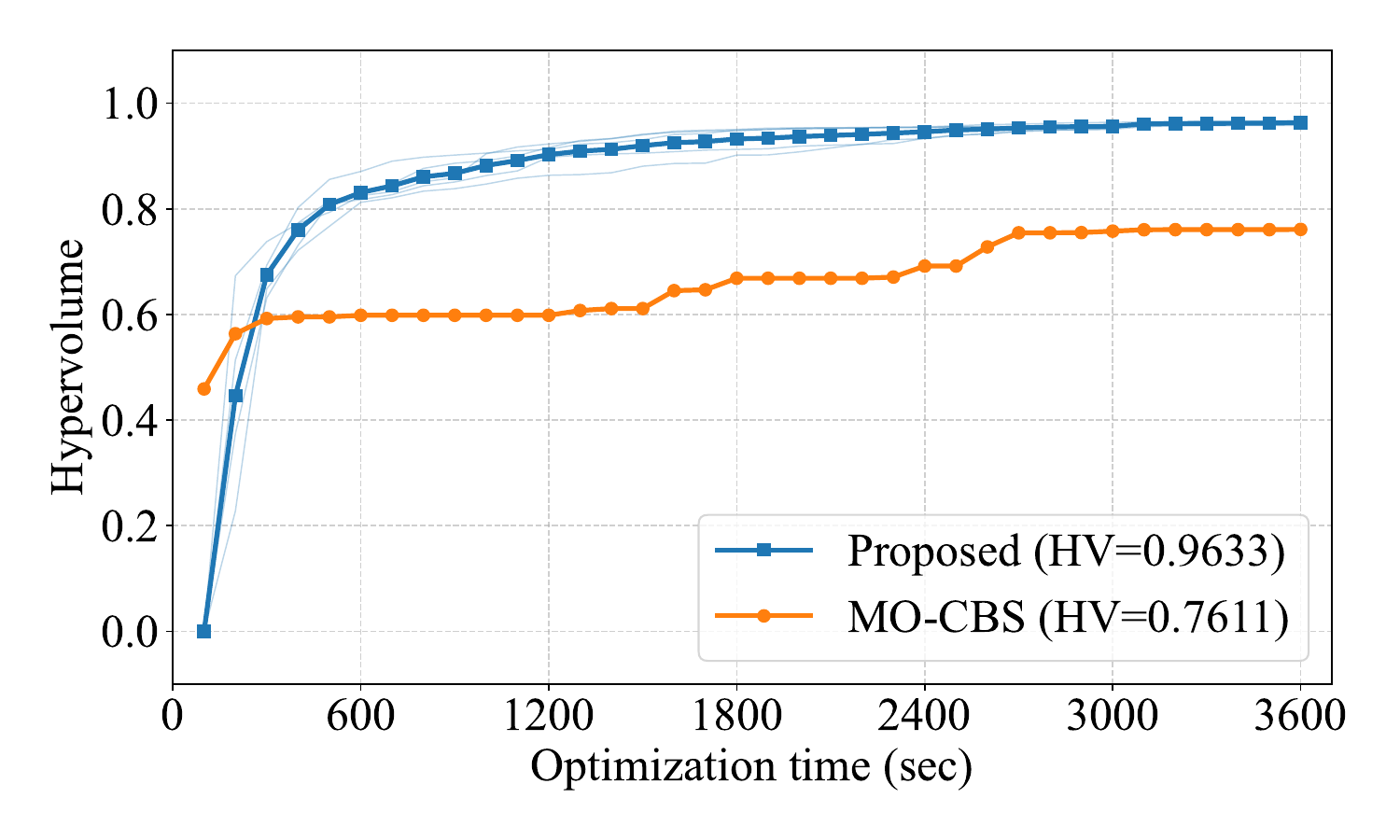}
\subcaption{G64, $N=15$}
\end{minipage}
\hfill
\begin{minipage}[b]{0.24\textwidth}
\centering
\includegraphics[width=\linewidth]{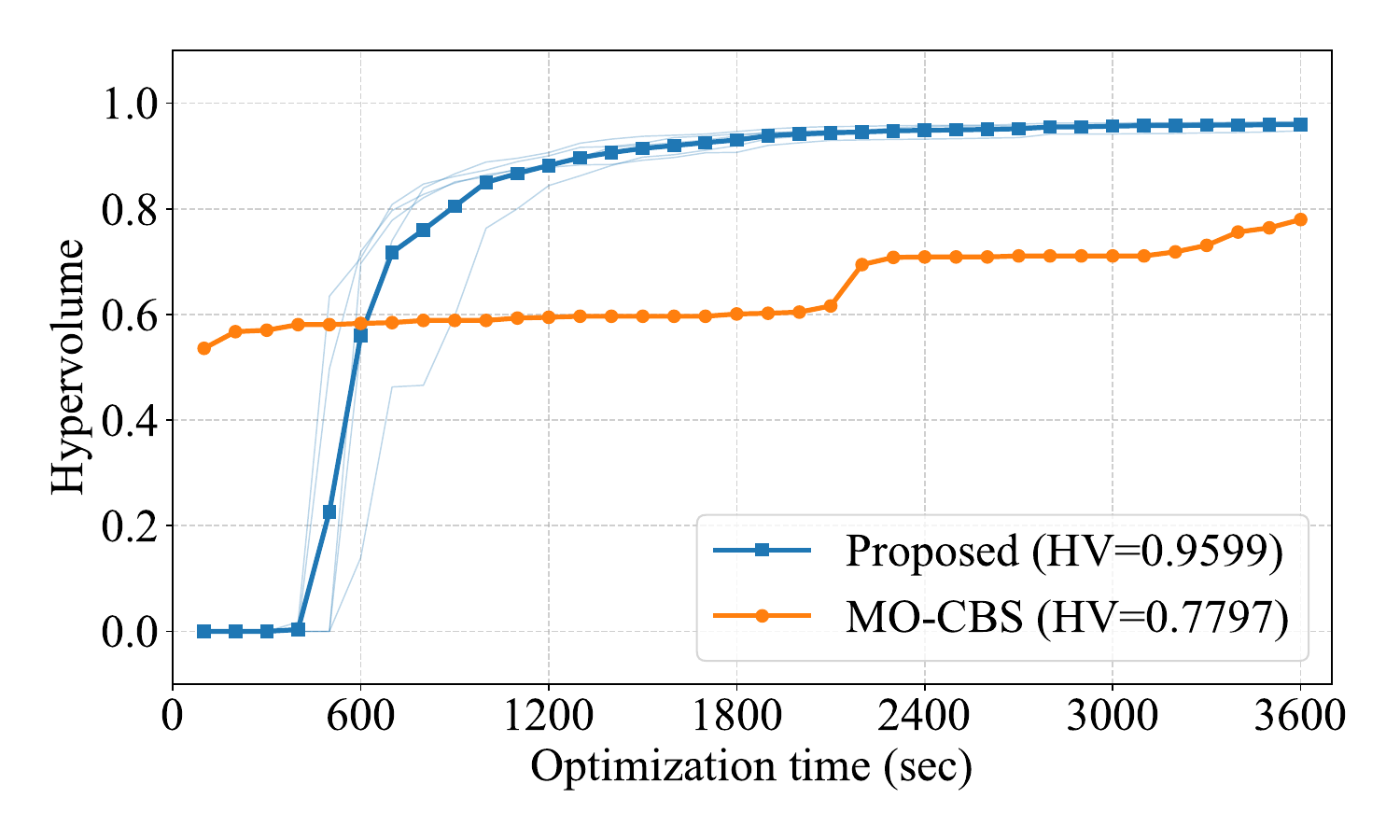}
\subcaption{G96, $N=10$}
\end{minipage}
\caption{Transition of HV calculated at 100-second intervals}
\label{fig_res_hv}
\end{figure}

Fig.~\ref{fig_res_hv} shows the transitions of the HV values for the obtained non-dominated solution sets, calculated at 100-second intervals.
The numerical values in the legend indicate the HV values for each method at 3,600 seconds.
For the proposed method, the HV values obtained in each trial are represented by thin lines, while their average is shown as a bold line. 

As shown in Fig.~\ref{fig_res_hv}, MO-CBS terminates within 100 seconds and finds POS for the small-scale G32 map. 
Throughout the entire 3,600-second time horizon, MO-CBS consistently maintains a higher HV value than the proposed method. 

In contrast, for the larger G64 and G96 maps, the proposed method surpasses MO-CBS in HV early in the search and outperforms it thereafter. 
Although MO-CBS initially achieves relatively high HV values, both algorithms exhibit HV values that are lower than their final results in the early stage, as the search space has not yet been sufficiently explored.

Furthermore, the proposed method attains HV values comparable to its final performance from the early phases of optimization. 
These results indicate that the proposed method delivers solutions with stable and consistent quality throughout the search process.

\begin{figure}[!tb]
\centering
\begin{minipage}[b]{0.24\textwidth}
\centering
\includegraphics[width=\linewidth]{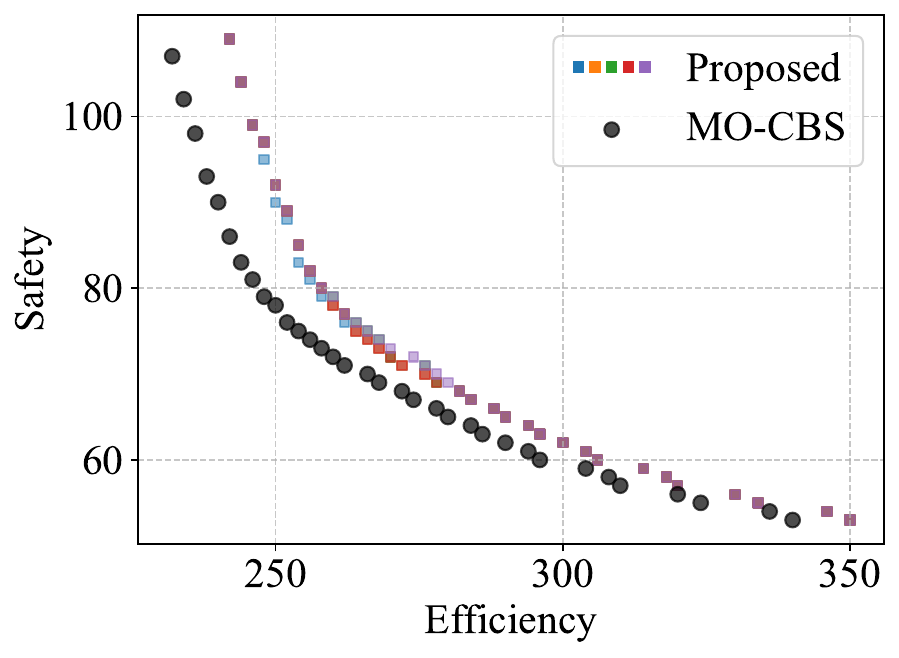}
\subcaption{G32, $N=10$}
\end{minipage}
\hfill
\begin{minipage}[b]{0.24\textwidth}
\centering
\includegraphics[width=\linewidth]{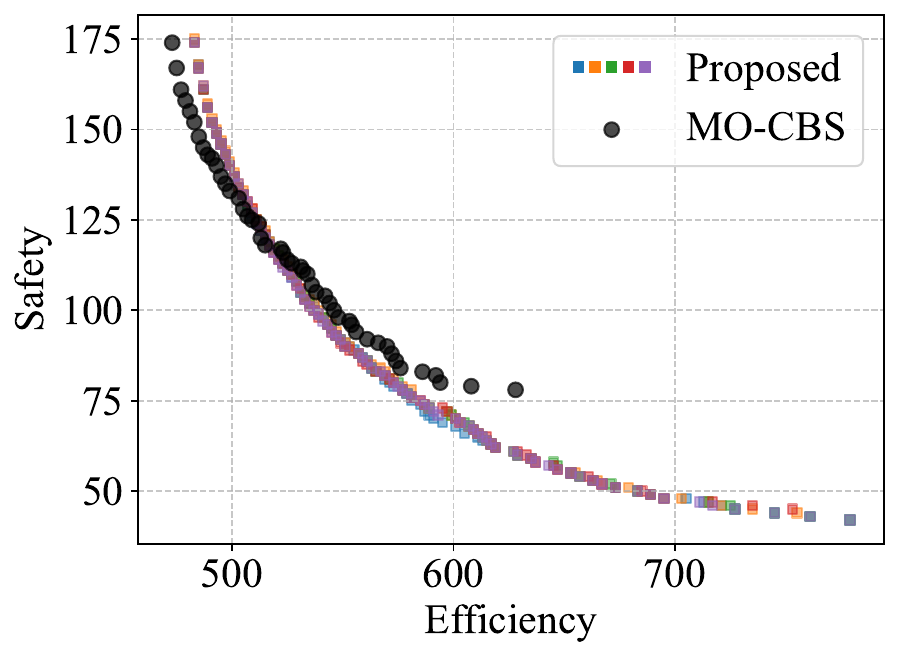}
\subcaption{G64, $N=10$}
\label{fig_G64_N10}
\end{minipage}
\hfill
\begin{minipage}[b]{0.24\textwidth}
\centering
\includegraphics[width=\linewidth]{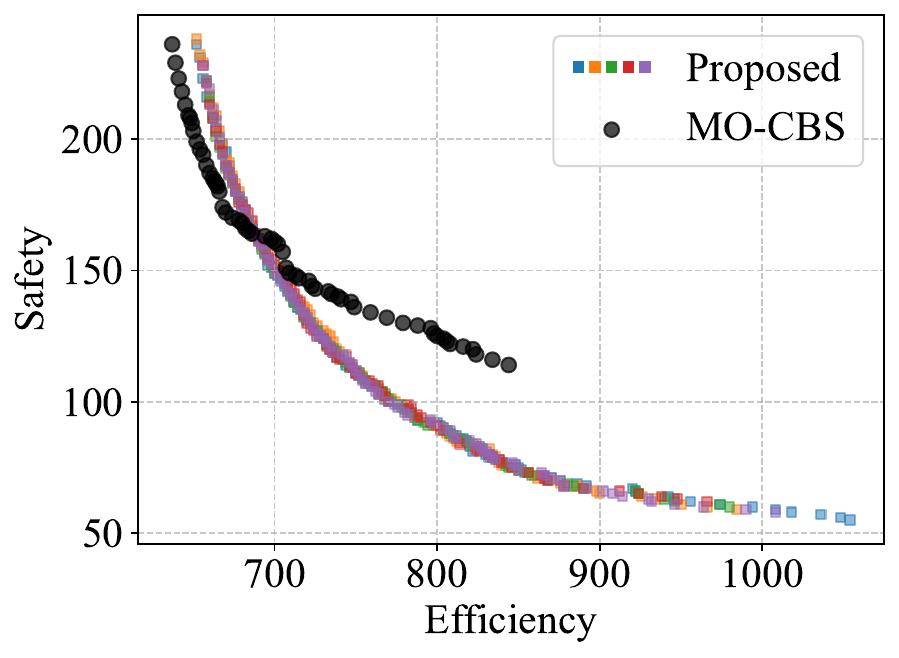}
\subcaption{G64, $N=15$}
\end{minipage}
\hfill
\begin{minipage}[b]{0.24\textwidth}
\centering
\includegraphics[width=\linewidth]{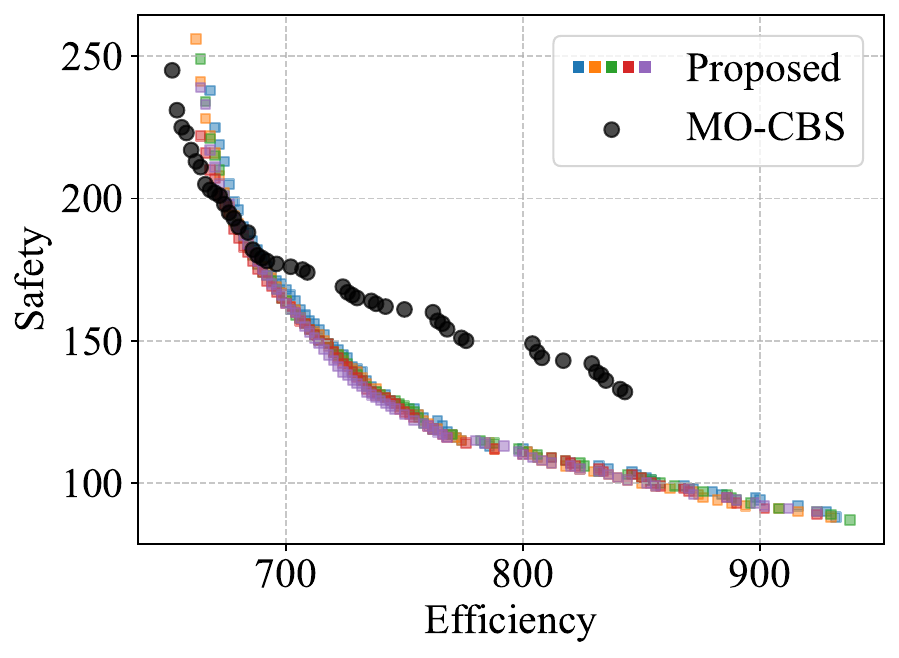}
\subcaption{G96, $N=10$}
\end{minipage}
\caption{Non-dominated solution set obtained by two methods}
\label{fig_res_obj}
\end{figure}
Fig.~\ref{fig_res_obj} shows the final non-dominated solution sets. 
The horizontal axis indicates the path length (Efficiency), and the vertical axis indicates the path cost (Safety). 
The results of the proposed method from five trials are shown as colored points, while those of MO-CBS are shown as black points.

As shown in Fig.~\ref{fig_res_obj}, MO-CBS finds a better Pareto front than the proposed method in G32.
On the other hand, MO-CBS failed to obtain the entire Pareto front in G64 and G96.
In contrast, the proposed method finds solutions over a wider range of the objective space than MO-CBS does in G64 and G96. 
In particular, the proposed method finds safer but less efficient paths in the lower-right region that MO-CBS does not find. 
This result stems from the fact that MO-CBS performs the search in lexicographical order, which causes optimization to proceed from the region where the path length is minimized. 
This difference becomes more pronounced as the problem scale increases. 
In G64, the difference is larger for $N=15$ than for $N=10$, and, for the same number of agents, it is larger in G96 than in G64. 
These results demonstrate the high scalability of the proposed method for large-scale problems.

\begin{figure*}[!tb]
\centering
\begin{minipage}[t]{0.24\linewidth}
\vspace{0pt}
\centering
\includegraphics[width=\linewidth]{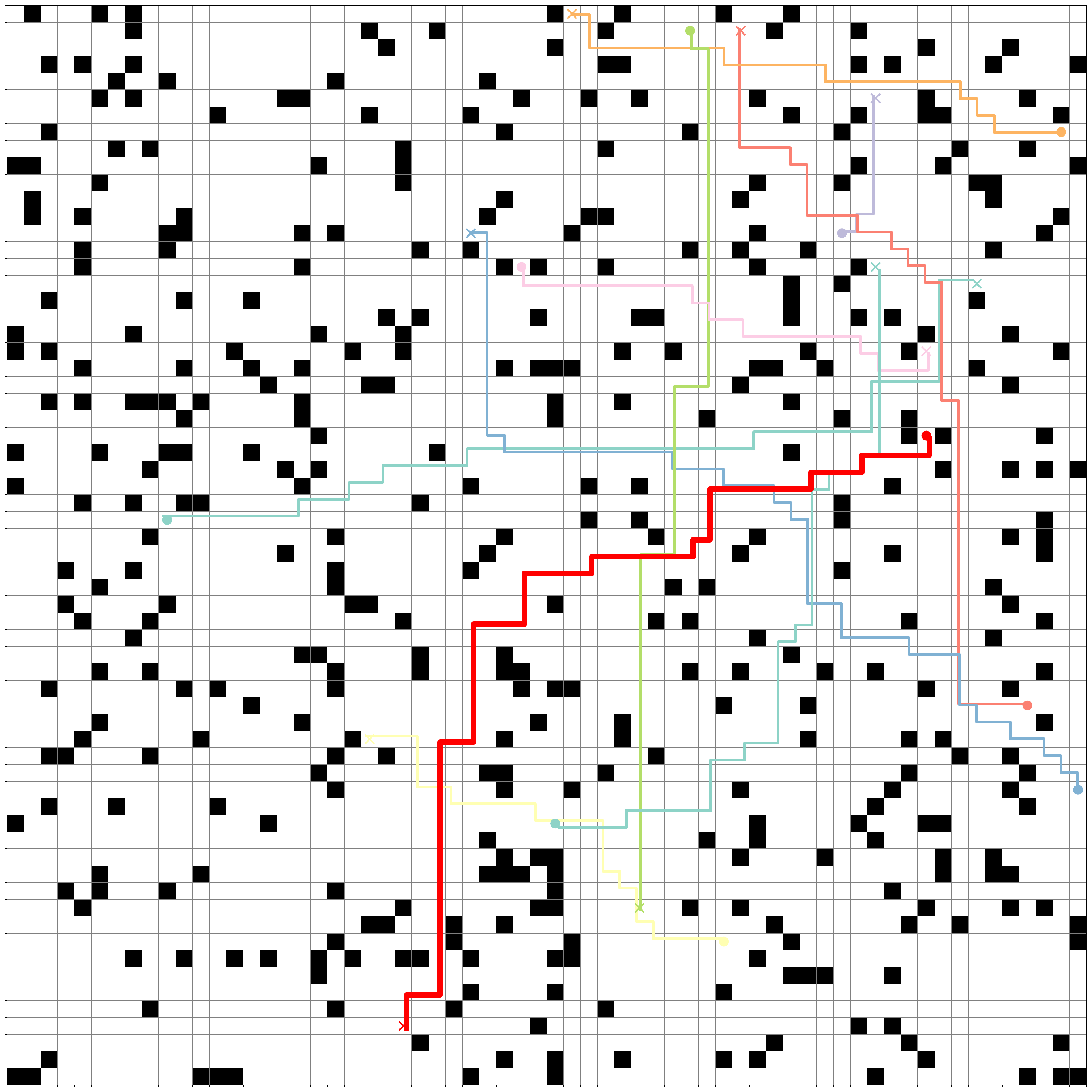}
\subcaption{Minimum path length solution\\(Efficiently: 66, Safety: 20)}
\label{fig_mlpath}
\end{minipage}
\hfill
\begin{minipage}[t]{0.24\linewidth}
\vspace{0pt}
\centering
\includegraphics[width=\linewidth]{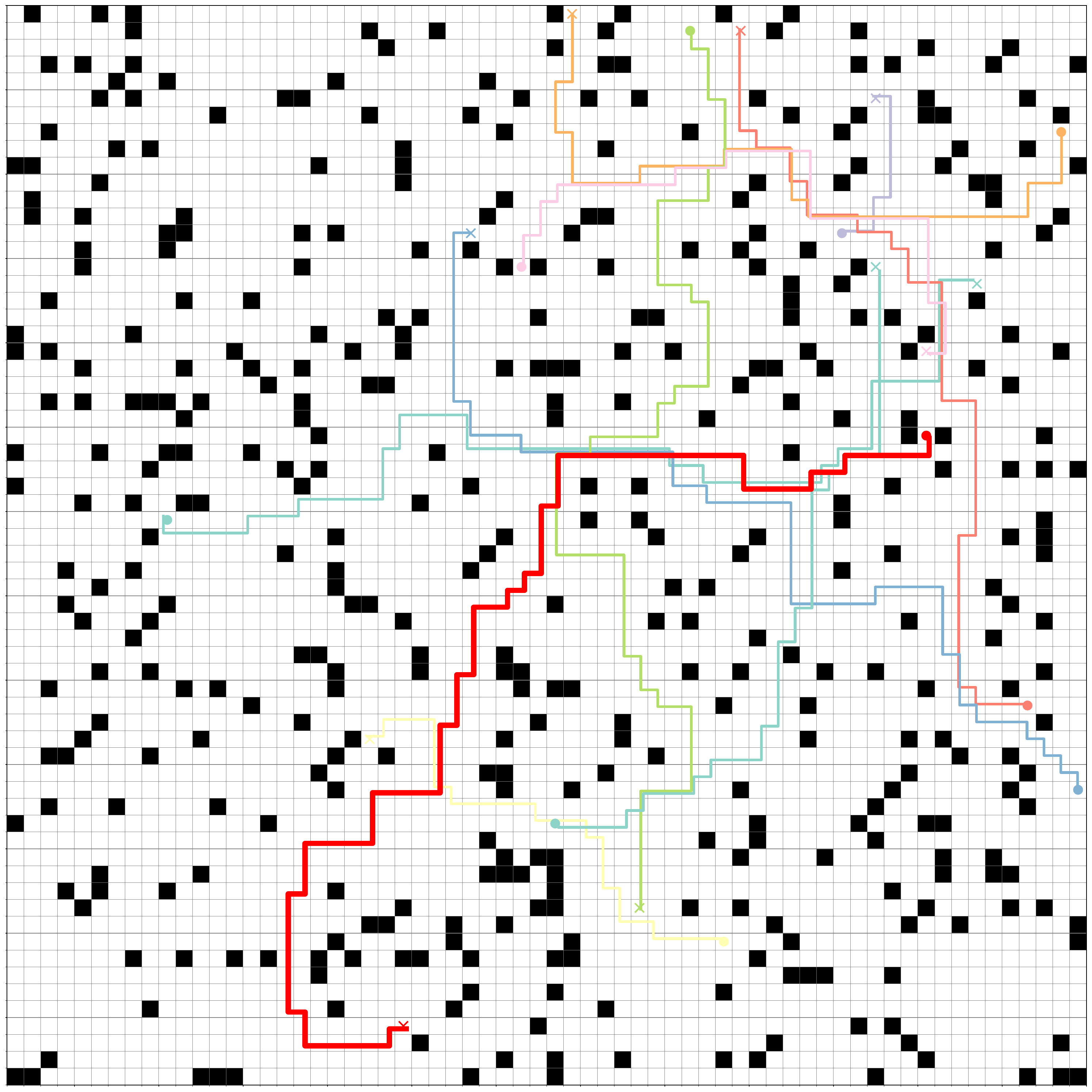}
\subcaption{Intermediate solution\\(Efficiently: 86, Safety: 10)}
\label{fig_inpath}
\end{minipage}
\hfill
\begin{minipage}[t]{0.24\linewidth}
\vspace{0pt}
\centering
\includegraphics[width=\linewidth]{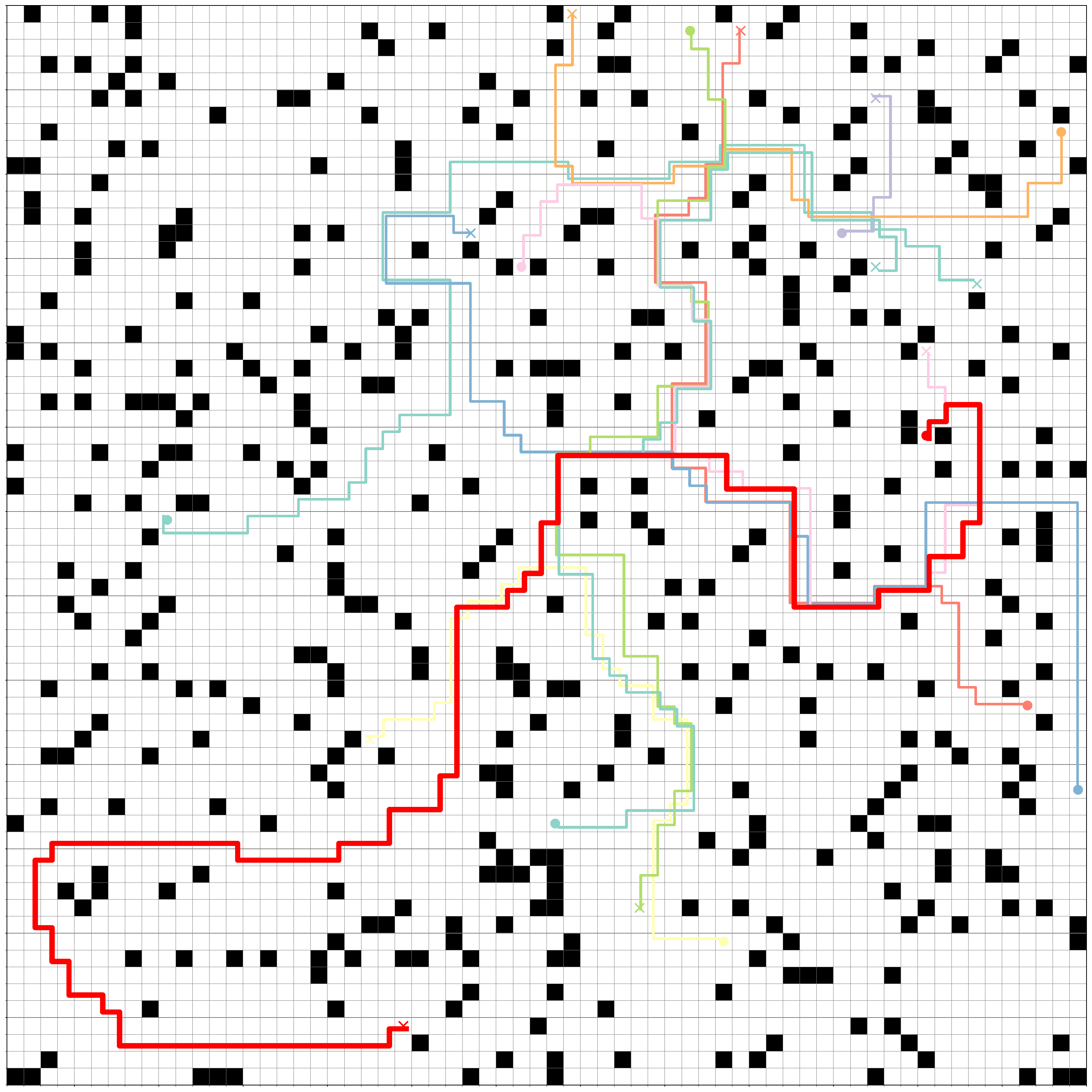}
\subcaption{Minimum path cost solution\\(Efficiently: 142, Safety: 5)}
\label{fig_mcpath}
\end{minipage}
\caption{Example of path sets obtained by the proposed method in G64 and $N=10$}
\label{fig_res_path}
\end{figure*}

Fig.~\ref{fig_res_path} shows the paths of each agent for three solutions obtained by the proposed method in a scenario with the G64 map and $N=10$ agents (Fig.~\ref{fig_G64_N10}): specifically, the solutions with the minimum path length, the minimum path cost, and an intermediate solution are illustrated, respectively.
In Fig.~\ref{fig_res_path}, black cells indicate the obstacles.

From Fig.~\ref{fig_res_path}, we see that the proposed method finds paths for each objective. 
In the solution where path length is minimized (Fig.~\ref{fig_mlpath}), agents follow the shortest path even if it involves grids near high-cost obstacles. 
In contrast, in the solution where path cost is minimized (Fig.~\ref{fig_mcpath}), agents take routes that detour around high-cost grids. 
For example, the agent indicated by the red line in Fig.~\ref{fig_mlpath} follows a linear path by accepting an increase in path cost when the path length is minimized. 
Conversely, when the path cost is minimized, the agent indicated by the red line in Fig.~\ref{fig_mcpath} follows a significantly longer detouring route.
The intermediate solution (Fig.~\ref{fig_inpath}) lies between the two. 
As a result, the balance between path length and path cost varies across solutions.

\section{Discussion}
Experimental results revealed that the proposed method achieves high performance in obtaining a comprehensive Pareto front within a fixed runtime, even for large-scale problems. 
The factors contributing to these results are discussed from two perspectives: scalability and solution diversity.

\subsection{High Scalability}
MO-CBS showed a tendency for solution quality to decrease compared to the proposed method as the problem size increased. 
This is due to the structure of MO-CBS, which explicitly detects and resolves conflicts among agents during high-level search. 
Specifically, as the number of agents $N$ increases, the number of conflicts to resolve grows, leading to a significantly larger search tree. 

In contrast, the proposed method is based on EAs and probabilistically searches for promising regions. 
It also includes processes to reduce the problem scale, such as limiting the search range during initialization and removing infeasible solutions that contain conflicts. 
These structural differences make the algorithm computationally efficient even as the problem scale grows, contributing to high scalability.

\subsection{Diversity of Solutions}
The solution set $\mathcal{S}$ produced by the proposed method yields a more comprehensive Pareto front in the search space than MO-CBS. 
This result stems from the fact that MO-CBS searches in lexicographical order of costs, leading to biased solutions during execution and preventing the algorithm from covering the entire Pareto front. 
This bias in MO-CBS is evident in the results for G64 and G96 in Fig.~\ref{fig_res_obj}. 

In contrast, the proposed method is based on EAs and explores the entire objective space simultaneously through a population-based search. 
As a result, it maintains a set of solutions that cover the entire Pareto front at any stage of the algorithm, allowing it to provide solutions across a wide range of the objective space within any given time. 
This difference helps ensure solution diversity within a limited running time.

\section{Conclusion}
This study proposed a novel method, TSE-MOMAPF, for solving large-scale MOMAPF problems within a realistic time frame. 
Experimental results demonstrated that the proposed method obtains higher-quality Pareto solutions under time constraints than MO-CBS. 
Specifically, the proposed method demonstrated strong scalability for large-scale problems, such as those with more agents or a larger map.

Future work will involve improving the proposed method to further enhance scalability. 
For example, while the current approach removes infeasible solutions that contain conflicts, we will investigate techniques to utilize these solutions to reach high-quality feasible solutions.

\section*{Acknowledgements}
During the preparation of this work, the authors used AI systems to improve language and readability. After using these tools, the authors reviewed and edited the content as needed and took full responsibility for the publication.

\bibliographystyle{IEEEtran}
\bibliography{reference}

@article{Ren_2023,
  author={Ren, Zhongqiang and Rathinam, Sivakumar and Choset, Howie},
  journal={IEEE Transactions on Automation Science and Engineering}, 
  title={A Conflict-Based Search Framework for Multiobjective Multiagent Path Finding}, 
  year={2023},
  volume={20},
  number={2},
  pages={1262-1274},
  doi={10.1109/TASE.2022.3183183}
}

@article{Stern_2021, 
  title={Multi-Agent Pathfinding: Definitions, Variants, and Benchmarks}, 
  volume={10}, 
  DOI={10.1609/socs.v10i1.18510}, 
  number={1}, 
  journal={Proceedings of the International Symposium on Combinatorial Search}, 
  author={Stern, Roni and Sturtevant, Nathan and Felner, Ariel and Koenig, Sven and Ma, Hang and Walker, Thayne and Li, Jiaoyang and Atzmon, Dor and Cohen, Liron and Kumar, T. K. and Barták, Roman and Boyarski, Eli}, 
  year={2021}, 
  month={Sep.}, 
  pages={151-158}
}

@article{sharon_2015,
  title={Conflict-based search for optimal multi-agent pathfinding},
  author={Sharon, Guni and Stern, Roni and Felner, Ariel and Sturtevant, Nathan R},
  journal={Artificial intelligence},
  volume={219},
  pages={40--66},
  year={2015},
  publisher={Elsevier}
}

@article{pulido_2015,
  title={Dimensionality reduction in multiobjective shortest path search},
  author={Pulido, Francisco-Javier and Mandow, Lawrence and P{\'e}rez-de-la-Cruz, Jos{\'e}-Luis},
  journal={Computers \& Operations Research},
  volume={64},
  pages={60--70},
  year={2015},
  publisher={Elsevier}
}

@inproceedings{ulloa_2020,
  title={A simple and fast bi-objective search algorithm},
  author={Ulloa, Carlos Hern{\'a}ndez and Yeoh, William and Baier, Jorge A and Zhang, Han and Suazo, Luis and Koenig, Sven},
  booktitle={Proceedings of the International Conference on Automated Planning and Scheduling},
  volume={30},
  pages={143--151},
  year={2020}
}

@article{eppstein_1998,
  title={Finding the k shortest paths},
  author={Eppstein, David},
  journal={SIAM Journal on computing},
  volume={28},
  number={2},
  pages={652--673},
  year={1998},
  publisher={SIAM}
}

@article{beume_2007,
  title={SMS-EMOA: Multiobjective selection based on dominated hypervolume},
  author={Beume, Nicola and Naujoks, Boris and Emmerich, Michael},
  journal={European journal of operational research},
  volume={181},
  number={3},
  pages={1653--1669},
  year={2007},
  publisher={Elsevier}
}

@inproceedings{weise_2020,
  title={On the scalable multi-objective multi-agent pathfinding problem},
  author={Weise, Jens and Mai, Sebastian and Zille, Heiner and Mostaghim, Sanaz},
  booktitle={2020 IEEE Congress on Evolutionary Computation (CEC)},
  pages={1--8},
  year={2020},
  organization={IEEE}
}

@ARTICLE{Guizzo_2008,
  author={Guizzo, Eric},
  journal={IEEE Spectrum}, 
  title={Three Engineers, Hundreds of Robots, One Warehouse}, 
  year={2008},
  volume={45},
  number={7},
  pages={26-34},
  doi={10.1109/MSPEC.2008.4547508}
}

@article{dresner_2008,
  title={A multiagent approach to autonomous intersection management},
  author={Dresner, Kurt and Stone, Peter},
  journal={Journal of artificial intelligence research},
  volume={31},
  pages={591--656},
  year={2008}
}

@ARTICLE{Alkazzi_2024,
  author={Alkazzi, Jean-Marc and Okumura, Keisuke},
  journal={IEEE Access}, 
  title={A Comprehensive Review on Leveraging Machine Learning for Multi-Agent Path Finding}, 
  year={2024},
  volume={12},
  number={},
  pages={57390-57409},
  doi={10.1109/ACCESS.2024.3392305}
}

@ARTICLE{Sharma2022-qi,
  title     = "A comprehensive review on multi-objective optimization
               techniques: Past, present and future",
  author    = "Sharma, Shubhkirti and Kumar, Vijay",
  journal   = "Arch. Comput. Methods Eng.",
  publisher = "Springer Science and Business Media LLC",
  volume    =  29,
  number    =  7,
  pages     = "5605--5633",
  month     =  nov,
  year      =  2022,
  language  = "en"
}

\end{document}